\documentclass[lettersize,journal]{IEEEtran}
\usepackage{amsmath,amsfonts}
\usepackage{array}
\usepackage[caption=false,font=normalsize,labelfont=sf,textfont=sf]{subfig}
\usepackage{textcomp}
\usepackage{stfloats}
\usepackage{url}
\usepackage{verbatim}
\usepackage{graphicx}
\usepackage{cite}
\usepackage{booktabs}
\usepackage{algorithm2e}
\usepackage{xcolor}
\usepackage{multirow}
\usepackage{setspace}
\begin{document}

\title{OccluTrack: Rethinking Awareness of Occlusion for Enhancing Multiple Pedestrian Tracking}

\author{IEEE Publication Technology,~\IEEEmembership{Staff,~IEEE,}
\author{Jianjun Gao, 
        Yi Wang, ~\IEEEmembership{Member,~IEEE,}
        Kim-Hui Yap, ~\IEEEmembership{Senior Member,~IEEE,}
        Kratika Garg, 
        and Boon Siew Han
    }

\thanks{This paper was produced by the IEEE Publication Technology Group. They are in Piscataway, NJ.}
\thanks{Manuscript received April 19, 2021; revised August 16, 2021.}}

\markboth{Journal of \LaTeX\ Class Files,~Vol.~14, No.~8, August~2021}%
{Shell \MakeLowercase{\textit{et al.}}: A Sample Article Using IEEEtran.cls for IEEE Journals}

\maketitle

\begin{abstract}

Multiple pedestrian tracking is crucial for enhancing safety and efficiency in intelligent transport and autonomous driving systems by predicting movements and enabling adaptive decision-making in dynamic environments. It optimizes traffic flow, facilitates human interaction, and ensures compliance with regulations. However, it faces the challenge of tracking pedestrians in the presence of occlusion. Existing methods overlook effects caused by abnormal detections during partial occlusion. Subsequently, these abnormal detections can lead to inaccurate motion estimation, unreliable appearance features, and unfair association. To address these issues, we propose an adaptive occlusion-aware multiple pedestrian tracker, OccluTrack, to mitigate the effects caused by partial occlusion. Specifically, we first introduce a plug-and-play abnormal motion suppression mechanism into the Kalman Filter to adaptively detect and suppress outlier motions caused by partial occlusion. Second, we develop a pose-guided re-identification (Re-ID) module to extract discriminative part features for partially occluded pedestrians. Last, we develop a new occlusion-aware association method towards fair Intersection over Union (IoU) and appearance embedding distance measurement for occluded pedestrians. Extensive evaluation results demonstrate that our method outperforms state-of-the-art methods on MOTChallenge and DanceTrack datasets. Particularly, the performance improvements on IDF1 and ID Switches, as well as visualized results, demonstrate the effectiveness of our method in multiple pedestrian tracking.

\end{abstract}

\begin{IEEEkeywords}
Multiple pedestrian tracking, tracking by detection, Kalman Filter, re-identification, data association
\end{IEEEkeywords}

\section{Introduction}

\IEEEPARstart{M}{ultiple} pedestrian tracking is indispensable for intelligent transport \cite{surveillance1} and autonomous driving systems \cite{self_driving, drive1} as it enables predicting pedestrian movements \cite{path_prediction}, adaptive decision-making \cite{review_tracking, dm_tro}, and compliance with regulations \cite{survey} in dynamic urban environments. By accurately monitoring pedestrian behavior, this technology enhances safety by minimizing collision risks \cite{collision} and optimizing traffic flow \cite{traffic}. Additionally, it facilitates efficient navigation for autonomous vehicles through crowded areas and ensures smooth interactions between vehicles and pedestrians. This task remains challenging, aiming to form trajectories for detected pedestrians over time. This process involves detecting pedestrians and associating identical pedestrians from sequential frames.

Existing state-of-the-art multiple pedestrian tracking methods \cite{sort, fairmot, bytetrack, botsort} associate detected pedestrians by combining cues like motions and appearance features to address the occlusion problem. Kalman Filter, one of the motion estimators, is commonly used to provide motion cues by formulating multiple pedestrian trackers as a linear estimation problem, which recursively predicts and updates trajectories from noisy observations (detections) over frames. Re-ID modules extract appearance features from multiple frames to identify the same pedestrian. However, these methods, even with Re-ID, are insufficient to resolve occlusion problems because of inaccurate motion estimates, unreliable appearance features, and unfair association.

Recently, occlusion has drawn much attention in multiple pedestrian tracking, and various approaches were proposed to address this issue, e.g., tracking by attention \cite{trackformer, transtrack}, graph neural networks \cite{transmot}, self and cross-attention mechanism \cite{deep-tracklet, motiontrack}, and hierarchical feature extraction \cite{quo-vadis}. They address the occlusion problem by formulating motion patterns via advanced deep learning models in adjacent or multiple frames or understanding the context information in the scene. However, these methods require significant computation resources compared with the Kalman Filter, and they ignore the effects caused by partial occlusion.

\begin{figure*}[tp]
    \centering
    \includegraphics[width=\linewidth]{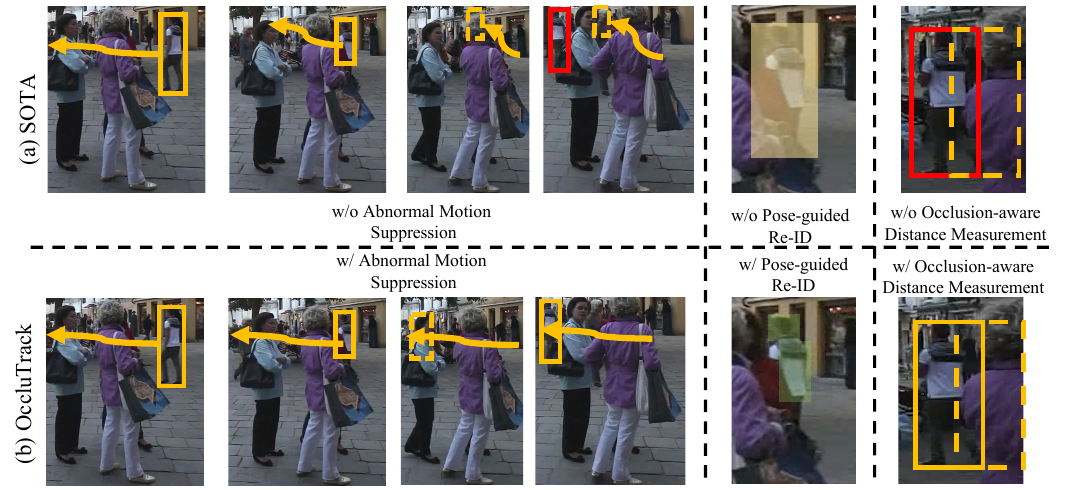}
    \caption{
The motivation of our proposed OccluTrack. (a) Left: the SOTA tracker \cite{botsort} struggles with abnormal motions caused by partial occlusion, leading to inaccurate motion estimation. Middle: They also lack pose guidance during partial occlusion, resulting in unreliable appearance features. Right: Treating occluded and non-occluded pedestrians equally in distance measurement hinders optimal association. (b) In contrast, OccluTrack incorporates abnormal motion suppression (left), pose-guided Re-ID (middle), and occlusion-aware distance measurement (right) to address these limitations.
    }
    \label{fig: Fig.1}
\end{figure*}

Through the visualization of experimental results, we found that partial occlusion is the missing key to resolving the occlusion problem. Partial occlusion creates abnormal bounding boxes covering some body parts only. These abnormal bounding boxes change suddenly in center points and aspect ratios, which causes inaccurate motion estimates, appearance features, and associations. As shown in Fig. \ref{fig: Fig.1}(a), abnormal detections caused by partial occlusion mislead the motion estimator in predicting wrong trajectories. Especially the errors will be accumulated and amplified without accurate observations during full occlusion. As for Re-ID modules, the inputs are from the cropped images according to the detections. Nevertheless, the cropped images will cover two persons (part of the obscured and part of the visible person) during partial occlusion, introducing ``unreliable and noisy appearance features''. Moreover, the predictions from the motion estimator during occlusion are not as accurate as those in normal situations because of the error accumulation caused by partial occlusion. Hence, the association method should not treat occluded persons as strictly as visible persons, i.e., ``unfair association''.

To alleviate the effects caused by partial occlusions in motion estimation, Re-ID, and association, we propose the OccluTrack, an adaptive occlusion-aware multiple pedestrian tracker with three strategies, as shown in Fig. \ref{fig: Fig.1}(b). First, we propose a plug-and-play abnormal motion suppression mechanism for stabilizing parameter updates in the Kalman Filter without effects on inference speed. In particular, the abnormal motion suppression mechanism leverages the history of tracked observations to detect and suppress abnormal motions. The motion can be more accurately predicted by abnormal motion suppression when partial occlusion occurs. Second, we introduce a pose-guided Re-ID strategy for robust part feature extraction. The pose-guided Re-ID strategy utilizes a pose estimator to guide the feature extractor to obtain more discriminative features. Lastly, we adopt an occlusion-aware distance measurement for occluded person association. The occlusion-aware distance measurement strategy involves a combination of IoU distance and appearance embedding distance under adaptive thresholds based on the level of occlusion, which is fair for occluded pedestrians.

We apply our method to three baseline trackers (JDE \cite{jde}, FairMOT \cite{fairmot}, and BoT-SORT\cite{botsort}), showing that the IDF1 and IDSw on the MOT17 validation set are improved consistently from 1.5\% to 2.4\% and 15\% to 30\%, respectively. We also evaluated our OccluTrack on the MOTChallenge \cite{mot17, mot20} and DanceTrack \cite{dancetrack} datasets, and the results show that our method outperforms other approaches by a significant margin under both private and public protocols.  
\begin{enumerate}
    \item We propose an abnormal suppression mechanism in the Kalman Filter to explicitly alleviate abnormal motions caused by partial occlusion. 
    \item We introduce a pose-guided Re-ID to extract more discriminative and reliable appearance features for partially occluded pedestrians. 
    \item We propose an occlusion-aware distance measurement method to make the association fair for occluded pedestrians.
    \item Extensive experiments were conducted on MOTChallenge and DanceTrack datasets, showing our proposed OccluTrack outperforms the state-of-the-art methods and has better tracking and association performance.
\end{enumerate}

\label{sec:intro}

\section{Related Work}
\subsection{Multiple Pedestrian Tracking}
At the early stage, works on multiple pedestrian tracking mainly focus on solving occlusion problems by motion cues. Intuitively, IoUTrack \cite{iou_tracker} associates the tracklets and detection solely based on the IoU. Only the IoU distance for detections from two adjacent frames indicates the geometric similarity. SORT \cite{sort} was the first work introducing Kalman Filter as the motion estimator into tracking-by-detection methods. Unlike IoUTrackers, SORT can predict the current positions upon the motion cues generated from Kalman Filter and enhance the tracking performance during occlusion. ByteTrack \cite{bytetrack} and OC-SORT \cite{ocsort} promoted the SORT by associating almost every detected object and smoothing the Kalman Filter by observations. Apart from SORT-based methods, CenterTrack \cite{centertrack} built on CenterNet \cite{centernet} adopts a regressor as the motion estimator, which regresses object motions from two adjacent frames. The following works \cite{ctracker, quasi} make CenterTrack more robust by introducing more proper motion formulation methods like incorporating past detections, cost volume, and Quasi-Dense similarities. STTD \cite{sttd} detects pedestrians with topologies and learns the dynamics of moving pedestrians within the same topology group.

Motion cues are not enough for occlusion avoidance because of a lack of accurate observations. The appearance feature \cite{zhou2022moving} is another cue to address the occlusion problem. DeepSort \cite{deepsort} first introduced Re-ID to solve the occlusion using appearance features. Following works \cite{fairmot, botsort, strongsort} adopt embedded or stand-by more robust Re-ID modules to enhance the ability to resolve the occlusion problem by modeling Re-ID more accurately. However, appearance features vary in different scenarios, especially during partial occlusion. 

Besides, some tracking-by-attention \cite{trackformer, metformer} methods modify DETR \cite{detr, deformabledetr} to an end-to-end tracker by introducing tracking queries and detection queries for tracked objects and new-born objects. These tracking-by-attention methods lack motion cues, which are sensitive to occlusions. Methods like P3AFormer \cite{p3aformer} re-introduce motion cues by methods like optical flows. Some other methods solve the occlusion problem from aspects of learning long-term temporal features and understanding scenes. DeepTracklet and MotionTrack \cite{deep-tracklet, motiontrack} make use of self- or cross-attention to learn long-term features. TransMOT \cite{transmot} and Quo Vadis\cite{quo-vadis} try to understand scenes from the perspective of graphs and Bird-Eye-View. However, these methods usually take dramatic computation resources and ignore the effects caused by partial occlusion.

\subsection{Motion Estimation with Kalman Filter}
Kalman Filter is commonly used in multiple pedestrian tracking with a data association algorithm such as the nearest neighbor and Hungarian algorithms. Data association algorithms associate the observations (e.g., object detections) with the predicted states, and the state estimates of the Kalman Filter are updated based on the associations. It is a recursive algorithm that estimates the state of a system based on noisy observations. It consists of two main steps: the prediction step and the update step. The two predictions are given by Eq. (\ref{eq:eq7}). At each time step $k$, the Kalman Filter first predicts the state of the system $\hat{s}_{k|k-1}$ using the previous state estimate $\hat{s}_{k-1|k-1}$. And matrix $A$ is the state-transition matrix that relates the state at time $k-1$ to the state at time $k$. Additionally, the Kalman Filter also predicts the covariance of the state estimate $P_{k|k-1}$ using the previous covariance estimate $P_{k-1|k-1}$ and a process noise $Q$ that describes how the uncertainty in the state evolves.

\begin{equation}
\label{eq:eq7}
    \text{Predict}: \begin{cases}
    \hat{s}_{k|k-1} = A \hat{s}_{k-1|k-1}\\
    P_{k|k-1} = A P_{k-1|k-1} A^\top + Q
    \end{cases}.
\end{equation}

After predicting the state and covariance estimates, the Kalman Filter updates them based on the current observation (or measurement) $z_k$, as formulated by Eq. (\ref{eq:eq8}). The observation model relates the observation to the state $z_k = H s_k + v_k$, where $H$ is the observation matrix that maps the state to the observation space, and $v_k$ is the observation noise with the covariance matrix of $R$.
The Kalman gain $G_k$ is first calculated in the update step, which determines the weight given to the observation. The updated state estimate $\hat{s}_{k|k}$ and covariance estimate $P_{k|k-1}$ are then computed.
$I$ is the identity matrix.
\begin{equation}
\label{eq:eq8}
    \text{Update}: \begin{cases}
    G_k = P_{k|k-1} H^\top (H P_{k|k-1} H^\top + R)^{-1} \\
    \hat{s}_{k|k} = \hat{s}_{k|k-1} + G_k (z_k - H \hat{s}_{k|k-1}) \\
    P_{k|k} = (I - G_k H) P_{k|k-1}
    \end{cases}.
\end{equation}

\subsection{Appearance Feature Extraction}
Appearance-based methods have proven to be effective in addressing occlusion challenges within the realm of multiple pedestrian tracking. Person re-identification (Re-ID) is a thoroughly researched problem in computer vision, with numerous studies proposing innovative algorithms and techniques to tackle it. In existing research \cite{cascade_transformer, pstr, partial}, appearance features are typically modeled using both global and local features. Global features encompass information extracted from the entire body, whereas local features are derived from various body parts. However, these conventional approaches may not adequately adapt to the demands of multiple pedestrian tracking, particularly in terms of efficiency and accounting for changes in appearance over time. In the domain of multiple pedestrian tracking, most existing methods primarily rely on global features, which can be susceptible to issues related to occlusion. For instance, DeepSORT \cite{deepsort} and subsequent methods such as BoT-SORT \cite{botsort}, StrongSORT \cite{strongsort}, and Deep OC-SORT \cite{deep_oc_sort} utilize a pre-existing Re-ID module to extract global features. Similarly, FairMOT \cite{fairmot} and its subsequent trackers \cite{ multi(TITS), interactively(TITS)} employ feature embeddings obtained from the detection feature map to represent global appearance. While some recent methods \cite{graph_attention_tracking, graph(TITS)} recognize the importance of incorporating local features to address occlusion challenges, they often employ complex network architectures like transformers, which may not be conducive to real-time tracking efficiency.
 
\subsection{Association for Occlusion}
The Hungarian algorithm plays a crucial role in linking detections and tracklets in tracking-by-detection methods by incorporating both motion and appearance distances. In the early stages of tracking-by-detection methods \cite{sort, centertrack}, the motion distance for association was typically computed using metrics such as Intersection over Union (IoU) or center distance. However, as the importance of appearance features became evident, the concept of appearance distance was introduced alongside motion distance to improve tracking performance, especially in scenarios involving occlusions. To optimize association in the presence of occlusions, various distance measurement strategies based on motion and appearance distances have been proposed. Most methods employ a weighted distance measurement approach \cite{association(TITS)}, where the two distances are combined using weighted averaging. More recently, approaches like BoT-SORT \cite{botsort} and Deep OC-SORT \cite{deep_oc_sort} have introduced innovative measurement methods. These methods first mask out the appearance distance using the motion distance and then select the minimum distance between them for the association. However, both weighted distance measurement and masked distance measurement tend to overlook the effects caused by partial occlusion. Partial occlusion can lead to unreliable predictions during occlusion periods. Consequently, applying the same distance measurement requirements for both occluded and visible pedestrians may not be equitable or optimal in addressing the challenges posed by partial occlusion.

\label{sec:related}

\section{Methodology}
\label{sec:method}
\subsection{Overview}
\begin{figure*}[tp]
    \centering
    \includegraphics[width=\linewidth]{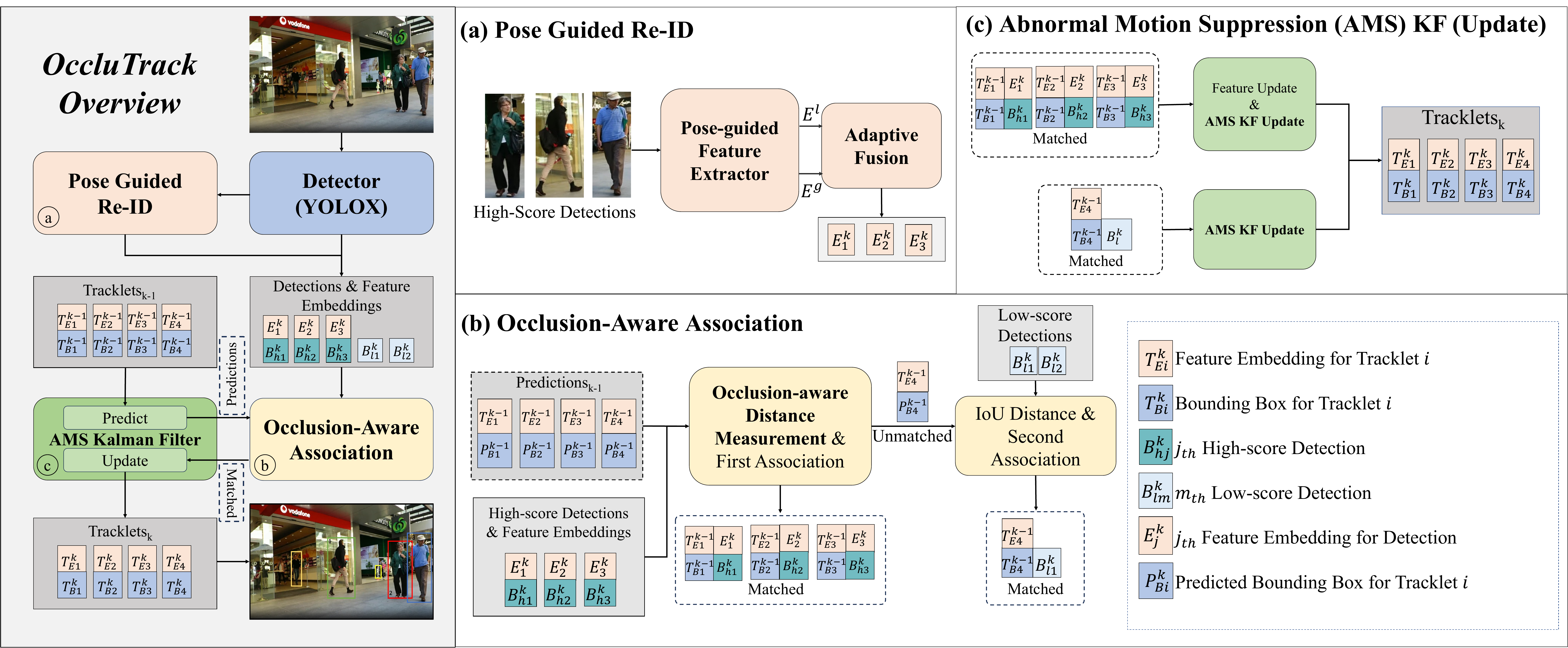}
    \caption{Overview of our proposed tracker. First, we utilize YOLOX \cite{yolox} as the object detector to obtain bounding boxes and separate them into high-score and low-score detections. For high-score detections, the pose-guided Re-ID module in (a) is used to extract appearance features. Second, the abnormal motion suppression Kalman Filter (AMS KF) is proposed to obtain $Predictions_{k-1}$ from previous $tracklets_{k-1}$. Based on the predictions, the association module with occlusion-aware distance measurement in (b) first associates the $Predictions_{k-1}$ of previous frames with the high-score detections (with feature embeddings), and then the second association matches unmatched predictions with low-score detections.  Finally, the abnormal motion suppression Kalman Filter in (c) is updated based on the matching results obtained from the association step. Specifically, the proposed AMS KF first detects the abnormal detections with abnormal changes. The $tracklets_{k}$ and parameters of the KF are updated correspondingly by considering the outlier detections caused by partial occlusion.}
    \label{fig: Fig.2}
\end{figure*}

Fig. \ref{fig: Fig.2} illustrates the overview of our proposed OccluTrack built upon the baseline tracker, BoT-SORT \cite{botsort}. It efficiently handles video frame processing for improved tracking in occlusion scenarios. In OccluTrack, it first performs object detections and separates detections into high-score and low-score detections. For high-score detections, we utilize our proposed pose-guided Re-ID module in Fig. \ref{fig: Fig.2}(a) to extract appearance features. Afterward, the abnormal motion suppression Kalman Filter (AMS KF) is used to provide predictions based on the tracklets from previous frames and pass the predictions to the occlusion-aware association module.  After that, detected pedestrian bounding boxes and appearance features are used to associate with detection and predictions from the Kalman Filter. Especially, the high-score detections are associated with our developed occlusion-aware distance measurement in Fig. \ref{fig: Fig.2}(b). For low-score detections, we associate them with predictions that are not matched with high-score detections using IoU distance. And associated or matched detections are used to update our abnormal motion suppression Kalman Filter in Fig. \ref{fig: Fig.2}(c). During the update, our proposed AMS KF will detect abnormal motions and suppress them when updating AMS KF parameters.

We present the detailed tracking algorithm for our proposed OccluTrack, as shown in Algorithm \ref{alg:two}. The tracking process is enhanced with specialized modules: pose-guided Re-ID (PG\_ReID), occlusion-aware distance measurement (ODM), and an abnormal motion suppression Kalman Filter (AMS\_KF). The algorithm starts by initializing tracklets with empty sets for bounding boxes, feature embeddings, and tracked bounding boxes buffer. For each frame in a video, pedestrian detection identifies bounding boxes, which are then categorized into high and low-confidence detections. High-confidence detections undergo feature embedding extraction via the PG\_ReID module. Kalman Filter predictions are made for tracklets, incorporating motion compensation. The distance matrix for high-confidence detections is calculated using ODM, and the Hungarian algorithm matches tracklets and detections. For low-confidence detections, matching is determined using the IoU distance metric, matched tracklets are updated with AMS\_KF parameters and the bounding box buffer. Tracklets that remain untracked beyond a threshold are discarded, and new tracklets are initiated for remaining detections with confidence above a certain threshold. This comprehensive algorithm optimizes tracking performance, particularly in challenging occlusion scenarios.

\subsection{Abnormal Motion Suppression Kalman Filter}

Kalman Filter has been widely adopted in tracking-by-detection methods, showing promising tracking performance. However, occlusion handling is a challenge within Kalman Filter. We observe that partial occlusion is one of the factors causing abnormal motions and wrong predictions during full occlusion, which is ignored in existing work. To mitigate the effects caused by partial occlusion, we introduce an abnormal motion suppression Kalman Filter to detect and suppress abnormal motions caused by partial occlusions during updating KF parameters.
\label{subsec:ms}
\paragraph{Partial Occlusion in Kalman Filter} In the update step of the Kalman Filter, we find that the observation $z_k$ at time step $k$ is incorrect during partial occlusion due to abnormal detections, i.e., objects’ bounding boxes suddenly become small due to partial occlusion. We assume that $z_k$ is corrupted by an error $e_k$, i.e., $z'_{k} = z_k + e_k$, where $z'_{k}$ is the corrupted observation. Hence, the updated state estimate in Eq.(\ref{eq:eq8}) becomes:
\begin{equation}
\label{eq:eq9}
    \tilde{s}_{k|k} = \hat{s}_{k|k-1} + G_k (z_k + e_k - H \hat{s}_{k|k-1}) = \hat{s}_{k|k} + G_k e_k,
\end{equation}
where $\tilde{s}_{k|k}$ is the corrupted state estimate disturbed by the $G_k e_k$, and $s=[x, y, w, h, \dot{x}, \dot{y}, \dot{w}, \dot{h}]^{\top}$ denotes the state vector, which includes the bounding box position $[x, y, w, h]^\top$ and its moving velocity  $[\dot{x}, \dot{y}, \dot{w}, \dot{h}]^\top$. During partial occlusion, the error $e'_k=G_k e_k=[e'_{xk}, e'_{yk}, e'_{wk}, e'_{hk}, e'_{\dot{x}k}, e'_{\dot{y}k}, e'_{\dot{w}k}, e'_{\dot{h}k}]^{\top}$ will dominate the update process. Subsequently, this error will accumulate during full occlusion without observations. In this period, the Kalman Filter's state update only relies on the predictions from the Kalman Filter. Here, we denote the duration for the full occlusion as $\tau$. As Eq. (\ref{eq:eq7}) indicates, the state is predicted by $\hat{s}_{k|k-1} = A \hat{s}_{k-1|k-1} $, where $A$ is the $n \times n$ transition matrix for the state from $k-1$ to $k$. Formally, $A$ can be elaborated for the linear motion as

\begin{small}
\begin{equation}
 \label{eq:eqs1}
 A = \begin{bmatrix}
  1 & 0 & 0 & 0 & 1 & 0 & 0 & 0 \\   
  0 & 1 & 0 & 0 & 0 & 1 & 0 & 0 \\ 
  0 & 0 & 1 & 0 & 0 & 0 & 1 & 0 \\ 
  0 & 0 & 0 & 1 & 0 & 0 & 0 & 1 \\ 
  0 & 0 & 0 & 0 & 1 & 0 & 0 & 0 \\ 
  0 & 0 & 0 & 0 & 0 & 1 & 0 & 0 \\ 
  0 & 0 & 0 & 0 & 0 & 0 & 1 & 0 \\ 
  0 & 0 & 0 & 0 & 0 & 0 & 0 & 1 \\ 
 \end{bmatrix}_{8 \times 8}.
\end{equation}
\end{small}

When the person reappears after occlusion of $\tau$ from $k0$, with only the prediction of the Kalman Filter, the estimate can be predicted like :
 \begin{equation}
 \label{eq:eqs2}
 \begin{split}
     \tilde{s}_{k0+\tau} &= A^{\tau} \tilde{s}_{k0} = A^{\tau} (\hat{s}_{k0} + G_{k0} e_{k0}) \\
     &= \hat{s}_{k0+\tau} + A^{\tau} G_{k0} e_{k0} = \hat{s}_{k0+\tau} + A^{\tau} e'_{k0}.
    \end{split} 
\end{equation}

According to the transition matrix $A$ and error $e'_k$, we can obtain $A^{\tau}$ and $e'_{k0}$. Hence, the accumulated error $A^{\tau} e'_{k0}$ can be calculated by 

\begin{scriptsize}
\begin{equation}
  \label{eq:eqs5}
 A^{\tau} e'_{k0} =\begin{bmatrix}
   1 & 0 & 0 & 0 & \tau & 0 & 0 & 0 \\   
   0 & 1 & 0 & 0 & 0 & \tau & 0 & 0 \\ 
   0 & 0 & 1 & 0 & 0 & 0 & \tau & 0 \\ 
   0 & 0 & 0 & 1 & 0 & 0 & 0 & \tau \\ 
   0 & 0 & 0 & 0 & 1 & 0 & 0 & 0 \\ 
   0 & 0 & 0 & 0 & 0 & 1 & 0 & 0 \\ 
   0 & 0 & 0 & 0 & 0 & 0 & 1 & 0 \\ 
   0 & 0 & 0 & 0 & 0 & 0 & 0 & 1 \\ 
  \end{bmatrix}  \begin{bmatrix}
   e'_{x0} \\   
   e'_{y0} \\ 
   e'_{w0} \\ 
   e'_{h0} \\ 
   e'_{\dot{x}0} \\ 
   e'_{\dot{y}0} \\ 
   e'_{\dot{w}0} \\ 
   e'_{\dot{h}0} \\ 
  \end{bmatrix} = \begin{bmatrix}
   e'_{x0}+\tau e'_{\dot{x}0} \\   
   e'_{y0}+\tau e'_{\dot{y}0} \\ 
   e'_{w0}+\tau e'_{\dot{w}0} \\ 
   e'_{h0}+\tau e'_{\dot{h}0} \\ 
   e'_{\dot{x}0} \\ 
   e'_{\dot{y}0} \\ 
   e'_{\dot{w}0} \\ 
   e'_{\dot{h}0} \\ 
  \end{bmatrix}
 \end{equation}
\end{scriptsize}

From Eq.(\ref{eq:eqs5}), we can observe that the bounding box $[x, y, w, h]^\top$ is shifted by a fixed error $[e'_{x0}, e'_{y0}, e'_{w0},e'_{h0}]^\top$ and an accumulated error $[\tau e'_{\dot{x}0},\tau e'_{\dot{y}0}, \tau e'_{\dot{w}0},\tau e'_{\dot{h}0}]^\top$ with a factor of $\tau$.

\RestyleAlgo{ruled}
\SetKwComment{Comment}{//}{}
\begin{algorithm}[hbt!] \footnotesize 
\caption{Pseudo-code of OccluTrack}\label{alg:two}

\SetKwInput{KwIn}{Input}
\SetKwInput{KwOut}{Output}

\KwIn{A video sequence $V$; object detecor \textit{Det}; Detection score threshold $\theta_{det}$; Pose-guided reID \textit{PG\_ReID}; Kalman Filter with Abnormal Motion Suppression \textit{AMS\_KF}; Occlusion-aware distance measurement \textit{ODM}; Threshold for initilizing new tracklets $\theta_{new\_track}$}
\KwOut{Tracklets $T = [T_B, T_E, BF_B]$ of the video}
\Comment{$T_B$:bounding boxes in tracklets, $T_E$:feature embeddings in tracklets,  $BF_B$:buffer for tracked bounding boxes}
Initialization: $T \gets [\emptyset, \emptyset, \emptyset]$\;
\For{$v$ in $V$}{
$B \gets Det(v)$ \;
$B_h \gets \emptyset$ \Comment*[l]{Initialize high-score bounding boxes}
$B_l \gets \emptyset$ \Comment*[l]{Initialize low-score bounding boxes}
\For{$b$ in $B$}{
    \eIf{$b.score>\theta_{det}$} {
        $B_h \gets B_h \cup \{b\} $\;
    }
    {
        $B_l \gets B_l \cup \{b\} $\;
    }
}
\Comment{\textcolor{green}{Pose-guided Re-ID}}

\For{$b$ in $B_h$}{
    $X = crop(v, b)$\;
    $E$ $\gets$ \textit{PG\_ReID} $(X, b)$\;
}

\Comment{Kalman filter prediction}

\For{$t_B$ in $T_B$}{
    $t_B$ = $KF\_AMS$.$predict(t_B)$ \;
    $t_B$ = \textit{MotionCompensation}($t_B, W$) \Comment*[l]{W:warp matrix from k-1 to k}
}
\Comment{\textcolor{green}{Occlusion-aware distance measurement for data association} }
$\hat{d}_h$ = \textit{ODM} $(T_B,T_E, B_h, E)$ \Comment*[l]{Associate $T$ and $B_h$ with $ODM$}
$matched_h, unmatched_h$ = \textit{Hungarian}$(\hat{d}_h)$ \;
$\hat{d}_l$ = \textit{IoU\_Distance}$(T_B[umatched_h[0,:]], B_l)$\Comment*[l]{Associate remained $T$ and $B_l$}

$matched_l, unmatched_l$ = \textit{Hungarian}$(\hat{d}_l)$ \;

\Comment{\textcolor{green}{Kalman filter update with abnormal motion suppression} }
\For{[$i$, $j$] in $matched_h$} {
\textit{AMS\_KF}.$update(T_B[i], B_h[j], BF_B[i])$ \; 
$T_E[i] = (1-\beta)*T_E[i]+\beta*E[j]$ 
}
\For{[$i$, $j$] in $matched_l$}{
    \textit{AMS\_KF}.$update(T_B[i], B_l[j], BF_B[j])$\;
}
$T \gets T \setminus T^{remain}$ \Comment*[l]{Remove remained tracklets $T^{remain}$ if untracked for a period}

\Comment{Initialize new tracks by remained detections $B^{remain}$ via $\theta_{new\_track}$}
\For{$b$ in $B^{remain}$}{
 \If{$b.score>\theta_{new\_track}$}{
    $T \gets T \cup \{b\}$\;
 }
}
}
Return \textbf{T}
\end{algorithm}

\paragraph{Kalman Filter Re-modeling} It is important to ensure that the observation error is properly modeled and accounted for in the Kalman Filter. To suppress abnormal motions during partial occlusion, we propose an abnormal motion suppression mechanism in Kalman Filter to detect and remove abnormal motions caused by partial occlusion. In particular, at Frame $k$, we first adopt a memory buffer to collect recent $N+1$ bounding boxes of a tracked pedestrian from $(k-N)$ to $k$, denoted as $\{B^{k}\}_{k=k-N}^{k}$ where $B^{k}=[x, y, w, h]^{\top}$ is a bounding box. Here, we apply a speed filter to calculate the averaged speed of the previous $(N-1)$ bounding boxes:
\begin{equation}
\label{eq:eq10}
    \bar{v} = \frac{1}{N-1} \sum_{n=k-N+1}^{k-1} (B^{n} - B^{n-1}),
\end{equation}
The current speed can be calculated by $v^{k} = B^{k} - B^{k-1}$, where $v=[v_x, v_y, v_w, v_h]^{\top}$. We define the speed of the center point and aspect ratio as $v_c=[v_x, v_y]^{\top}$ and $v_a=[v_w, v_h]^{\top}$, respectively. Then, we calculate the difference between the normalized current speed and the normalized average speed in the center point and aspect ratio, i.e., $d_c = \|v_{c}^{k}\| - \|\bar{v}_{c}\|$ and $ d_a = \|v_{a}^{k}\| - \|\bar{v}_{a}\|$. Finally, we obtain the speed difference for frame $k$ by $d^{k} = [d_c^{\top}; d_a^{\top}]^{\top} = [d_x, d_y, d_w, d_h]^{\top}$.

With $d^{k}$, we can set a threshold $\theta_v$ to detect the abnormal center point moving and scale changing of a tracked person. Afterward, we suppress the Kalman gain in the update step, Eq. (\ref{eq:eq8}), to handle the detected abnormal motions caused by the observation error ($e_k$) during partial occlusion. Formally, we calculate the suppression coefficients $\{\alpha_x, \alpha_y, \alpha_w, \alpha_h\}$ for each element of $d^{k}$ by:
\begin{equation}
\label{eq:eq13}
    \alpha_i = \begin{cases}
        1, \text{ if } d^{k}_{i} \leq \theta_v \\
       \alpha_0, \text{ otherwise}
    \end{cases},
\end{equation}
where $\alpha_0<1$ is a hyper-parameter to perform suppression. The average suppression coefficient $\alpha = (\alpha_x + \alpha_y + \alpha_w + \alpha_h)/4$ is used to scale the Kalman gain of the update step, so the updated state estimate in Eq. (\ref{eq:eq8}) is reformulated by:
\begin{equation}
\label{eq:eq15}
    \tilde{s}_{k|k} = \hat{s}_{k|k-1} + \alpha G_k (z_k - H \hat{s}_{k|k-1}) + \alpha G_k e_k.
\end{equation}

With the suppression coefficient, the model suppresses the state estimate by $\alpha<1$ when abnormal motions occur ($e_k>0$), more trusting the predicted state $\hat{s}_{k|k-1}$, and keeps the state estimate by $\alpha=1$ in normal motions ($e_k=0$).

While addressing the challenge of error accumulation within the Kalman Filter, OC-SORT \cite{ocsort} concentrates on updating the Kalman Filter solely when pedestrians can still be tracked after occlusion. This approach emphasizes incorporating velocity costs derived from previous observations into the association cost for updates. However, a notable drawback of this method is its tendency to overlook the impact of abnormal observations. By solely relying on observations, OC-SORT fails to account for the potential inaccuracies introduced by these abnormal observations, especially during instances of partial occlusion. This limitation becomes evident through the resulting significant disparities in velocities, leading to inaccurate association costs and, consequently, inaccurate associations.

In contrast, our proposed approach, OccluTrack, takes a distinct direction by focusing on mitigating the error accumulation issue during partial occlusion itself rather than merely after its occurrence. The central premise of OccluTrack is to rectify the adverse effects introduced by abnormal occlusions. This innovative approach aims to yield superior motion estimation accuracy during occlusion periods.

\subsection{Pose-Guided Re-ID}

\label{subsec:pg-reid}
\begin{figure*}[tp]
    \centering
    \includegraphics[width=0.9\linewidth]{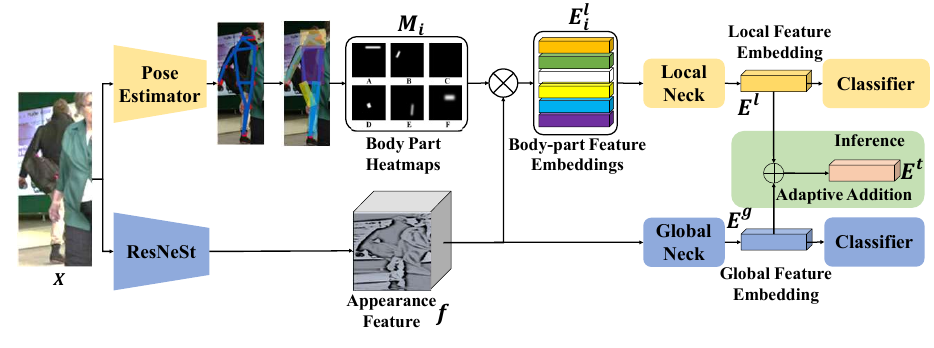}
    \caption{Architecture of our proposed pose-guided Re-ID module. It first extracts appearance features from the cropped image and estimates the pose of the occluded person. After that, body-part features are generated by applying body-part heatmaps to appearance features. By utilizing our proposed local and global necks, local and global feature embeddings are obtained for training. With the adaptive fusion, local and global embeddings are combined for inference. }
    \label{fig: Fig.3}
\end{figure*}

Re-ID has been introduced to mitigate the occlusion issue within tracking-by-detection methods. Existing methods extract the full-body features to calculate similarities for Re-ID. However, the full-body features get noisy when partial occlusion happens. To address this issue, we introduce a pose-guide Re-ID module to extract features by body parts rather than full-body.

\paragraph{Partial Occlusion in Re-ID} In multiple pedestrian tracking, Re-ID modules in prior works normally treated the bounding boxes equally using global features regardless of whether the pedestrians are obscured or not. They overlooked the significance of local features that contain more identity information, particularly during partial occlusion. Some methods \cite{tang2022weakly} employed Mask-RCNN \cite{maskrcnn} to extract local features, but they are coupled with detectors and challenging to optimize. In contrast, our approach introduces a pose-guided Re-ID module to adaptively fuse local and global features according to the confidence of body parts, generating more informative person appearance representations.

\begin{figure}
    \centering
    \includegraphics[width=0.4\textwidth]{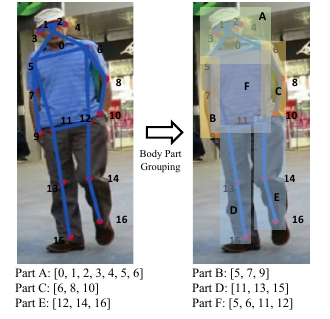}
    \caption{Combination of body keypoints to form different body parts. The keypoints are grouped into six body parts according to the given rules.}
    \label{fig: Fig.4}
\end{figure}

\paragraph{Pose-guided Feature Extractor} As depicted in Fig. \ref{fig: Fig.3}, we combine a real-time pose estimator with the Re-ID model to obtain and fuse local body-part and global feature embeddings. Given a high-score bounding box $B^k_h$, we crop the persons from the input image as the input for the Re-ID model.  From the figure, the lower-stream ResNeSt \cite{resnest} extracts the appearance feature $f \in \mathbb{R}^{H \times W \times C}$, where $H$ and $W$ denote the height and width of the feature map, and $C$ represents the channel size. Simultaneously, the upper-stream fast pose estimator \cite{alphapose} produces a heatmap $h \in \mathbb{R}^{H \times W \times 17}$ after Gaussian blur, indicating 17 keypoint positions and confidence scores. To reduce computation complexity, we group the heatmap $h \in \mathbb{R}^{H \times W \times 17}$ into the body-part heatmap $h' \in \mathbb{R}^{H \times W \times 6}$ by summation, as illustrated in Fig. \ref{fig: Fig.4}. Consequently, we use the heatmap to create six masks $hm_i \in \mathbb{R}^{H \times W\times 1}$, where $i \in [0,5]$. Each mask is repeated in channels, producing a mask of $M_i \in \mathbb{R}^{H \times W \times C}$.  Six body-part feature maps $f_i^l$ are obtained by $f_i^l = f \otimes M_i$, where $\otimes$ is the Hadamard product. The body-part feature embedding $E^l_i \in \mathbb{R}^{C}$ is generated by $E^l_i=\sum_{h=1}^{H}\sum_{w=1}^{W}f^l_i(h,w)$. 

We design local and global necks to extract the local and global feature embeddings. In the local neck, we concatenate the six body-part feature embeddings into $E^l_{cat}=[(E^l_0)^{\top}; (E^l_1)^{\top}; ...; (E^l_5)^{\top}]^{\top}$ and employ a fully-connected layer to project the channels from $6 \times C$ to $C$, resulting in a local feature embedding $E^l$. Conversely, the global neck applies the global average pooling for the appearance feature $f$, producing a global feature embedding $E^g \in \mathbb{R}^{C}$. During training, we employ two classifiers with batch normalization for the global and local feature embeddings. The Cross-Entropy loss and Triplet loss serve as the loss functions for each classifier, and we sum the four losses to form the overall loss during training:
\begin{equation}
\label{eq:eq16}
    \mathcal{L}  = \mathcal{L}^{ce}_l + \mathcal{L}^{triplet}_l + \mathcal{L}^{ce}_g + \mathcal{L}^{triplet}_g.
\end{equation}

\paragraph{Adaptive Fusion} During inference, feature embeddings associate tracklets appearance-wise by calculating the cosine similarity between the current feature embeddings and those from the previous tracklets. Thus, we should combine the local and global feature embeddings into one embedding before association. We propose an adaptive fusion method to combine local and global feature embeddings according to body-part confidence scores. The insight is to use the confidence scores of body parts to decide the weight of the local features, $n_p \in [0,6]$. The $n_p$ is obtained by counting the number of body parts whose confidence scores are higher than the threshold $\theta_p$.
The adaptive fusion combines two embeddings with the weight ($n_p$):
\begin{equation}
\label{eq:eq17}
    E^k = \frac{n_p}{6}*E^l + \frac{6-n_p}{6}*E^g.
\end{equation}

\subsection{Occlusion-aware Distance Measurement}

In data association, existing methods treat occluded and visible objects equally using the same association measurement. However, predictions for occluded objects may not be as accurate as those for visible objects, caused by factors like partial occlusion. Thus, in our occlusion-aware distance measurement, we loosen the requirements for associating objects that have been occluded.

\label{subsec:mea}
\paragraph{Partial Occlusion in Data Association} In state-of-the-art trackers, like BoT-SORT~\cite{botsort}, the IoU and appearance feature distances are performed conditionally. Specifically, the IoU and appearance embedding distances are first calculated between the high-score detections with feature embeddings and the tracklets. Conditioning on the IoU distance larger than a threshold, the final distance is chosen from the smaller one between the IoU distance and the appearance feature distance. However, the IoU distance is usually inaccurate because of error accumulation during partial occlusion, as mentioned in Section \ref{subsec:ms}.

\paragraph{Distance Measurement} To address the issue, we treat the observed tracklets and occluded tracklets in different manners. We adopt the occlusion-aware distance thresholds for two different types of tracklets. For observed tracklets, we keep the same settings as in BoT-SORT. For occluded tracklets, we set a higher IoU distance threshold ($\theta^{iou}_i$) and make it easier to keep track of occluded persons when the bounding boxes re-appear. In practice, we treat inactive tracklets as occluded tracklets as the tracklet will be set to inactive when it is occluded. Formally,
\begin{equation}
\label{eq:eq18}
    \theta^{iou}_i = \begin{cases}
    \theta_0 \text{ if } i \in I_k \\
    \theta_0+o \text{ if } i \in I_{uk}
    \end{cases},
\end{equation}
where $o>0$ is the offset for the distance threshold, $i$ is the index for all tracklets,  $I_k$ includes indices for tracked tracklets, $I_{uk}$ includes indices for untracked tracklets, and $\theta^{iou}_i$ represents the IoU distance threshold for Tracklet $i$. Then, we calculate the IoU distance $d^{iou}_{i,j}$ between Tracklet $i$ and Detection $j$ and the appearance embedding distance $\hat{d}^{cos}_{i,j}$ by:
\begin{equation}
\label{eq:eq19}
    \hat{d}^{cos}_{i,j} = \begin{cases}
    d^{cos}_{i,j}, \text{ if } d^{cos}_{i,j}<\theta^{emb} \cap d^{iou}_{i,j}<\theta^{iou}_i \\
    1, \text{ otherwise }
    \end{cases},
\end{equation}
where $\theta^{emb}$ is the proximity threshold of appearance embedding distance.  Finally, the occlusion-aware distance measurement is obtained by:
\begin{equation}
\label{eq:eq20}
    \hat{d}_{i,j} = \mathop {\min}(d^{iou}_{i,j}, \hat{d}^{cos}_{i,j}).
\end{equation}


\section{Experiments}

\subsection{Dataset}

We evaluate our proposed method on three tracking datasets: MOT17 \cite{mot17}, MOT20 \cite{mot20}, and DanceTrack \cite{dancetrack}. The MOT17 \cite{mot17} dataset consists of 7 training and 7 testing video sequences. The videos were captured in various real-world scenarios, such as outdoor scenes, crowded public places, and indoor environments. The dataset contains more than 22,000 annotated pedestrian trajectories, with varying levels of occlusion, illumination changes, and other challenging factors. The MOT20 \cite{mot20} dataset contains 4 training and 4 testing video sequences. The videos were captured in similar scenarios as MOT17 but with a larger crowd density. The dataset contains over 60,000 annotated object trajectories, with challenging scenarios such as crowdedness. DanceTrack \cite{dancetrack} is a benchmark designed to track multiple objects with uniform appearances and diverse motions. The dataset consists of 100 videos, with 40 videos for training, 25 for validation, and 35 for testing.

\subsection{Evaluation Metrics}

For all three datasets, we adopt CLEAR metrics \cite{clear}. The metrics include multiple object tracking accuracy (MOTA), IDF1 score, higher-order tracking accuracy (HOTA), and ID switch (IDSw). More importantly, MOTA quantifies the overall tracking accuracy by accounting for both false positives (FP) and false negatives (FN). IDF1 evaluates the tracking capability to maintain object identities consistently over time. And HOTA is a more balanced evaluation metric of detection and tracking. HOTA extends MOTA on assessment by incorporating higher-order tracking errors, including fragmentation and merging. IDSw quantifies the number of identity switches occurring throughout the video sequence.

\subsection{Experimental Details}

We adopt YOLOX as the object detector following the settings in our baseline, where YOLOX is trained on multiple datasets \cite{motchallenge, cityscapes, ethz, shao2018crowdhuman}. As for the parameters in our proposed method, we set the $\alpha_0 = 0.2$ in Eq. (\ref{eq:eq13}). And we set the buffer size of the bounding boxes the same as the buffer size for feature embeddings, which is 30. We adopt the FastPose in Alphapose \cite{alphapose} as the pose estimator to extract the heatmaps. We adopt ResNest \cite{resnest} as the appearance feature extractor in pose-guided Re-ID. And we follow the training strategies in our baseline for our pose-guided Re-ID module. We train the Re-ID model for 60 epochs based on the first half of the MOT17, MOT20, and DanceTrack training sets, respectively. During training, the pose estimator is frozen and only provides the heatmaps. In the occlusion-aware distance measurement, we set the offset $o = 0.2$ in Eq. (\ref{eq:eq18}). For other parameters, we keep the same settings in our baseline, where the $\theta^{cos} = 0.25$ and $\theta^{iou} = 0.5$. Before submitting results to MOTChallenge, we also perform interpolation on the trajectories following our baseline \cite{botsort}.

\subsection{Ablation Study}

\paragraph{Components Analysis}

\begin{table}[]
     \captionsetup{justification=justified, singlelinecheck=false, format=plain}
    \caption{The results of the components ablation study on three different existing trackers over the MOT17 validation set.  We applied these three modules (AMS KF, PG Re-ID, and OD Mea.) to three different existing trackers (JDE, FairMOT, and BoT-SORT) to demonstrate the effectiveness of our proposed method.
    }
    \label{table: ablation study}
    \centering
    \resizebox{0.5\textwidth}{!}{
    \begin{tabular}{lccccc}
    \toprule
        Tracker & AMS KF & PG Re-ID & OD Mea. & IDF1 (\%) ↑ & IDSw↓ \\
        \midrule
        JDE & ~ & ~ & ~ & 63.6 & 473 \\ 
        +AMS KF~ & \checkmark & ~ & ~  & 66.2 & 356 \\ 
        +PG Re-ID~ & \checkmark & \checkmark & ~ & \textcolor{blue}{66.6} & \textcolor{blue}{347} \\ 
        +OD Mea.~ & \checkmark & \checkmark & \checkmark & \textbf{66.9} & \textbf{327} \\
        \midrule
        FairMOT  & ~ & ~ & ~ & 72.8 & 299 \\ 
        +AMS KF~ & \checkmark & ~ & ~  & \textbf{74.9} & 268 \\ 
        + PG Re-ID~ & \checkmark & \checkmark & ~  & 74.7 & \textbf{223} \\ 
        +OD Mea.~ &  \checkmark & \checkmark & \checkmark  & \textcolor{blue}{74.8} & \textcolor{blue}{227} \\
        \midrule
        BoT-SORT & ~ & ~ & ~  & 81.5 & 147 \\ 
        +AMS KF~ & \checkmark & ~ & ~  & 82.3 & 134 \\ 
        +PG Re-ID~ & \checkmark & \checkmark & ~  & \textcolor{blue}{82.7} & \textcolor{blue}{129} \\ 
        +OD Mea.~ & \checkmark & \checkmark & \checkmark & \textbf{83.0} & \textbf{125} \\
        \bottomrule
    \end{tabular}
    }
\end{table}

We assess the impacts of our proposed components on tracking performance of three existing trackers: JDE, FairMOT, and BoT-SORT. We conduct our experiments on MOT 17 validation set following the strategies in ByteTrack \cite{bytetrack}. By gradually incorporating the three components, we evaluate the improvements achieved by each component cumulatively. The results of our ablation study are summarized in Table \ref{table: ablation study}, where we bold the best results and highlight the second best in blue.

To begin, we replace the original Kalman Filter in all three trackers with our abnormal motion suppression Kalman Filter. The results demonstrate notable performance gains compared to the original trackers. Applying abnormal motion suppression Kalman Filter to JDE, the IDF1 improves by 2.6\% compared to the original JDE, and the IDSw decreases dramatically from 473 to 356. FairMOT also benefits from our proposed Kalman Filter, with IDF1 improvements of 2.1\% and a drop in IDSw of 31, achieving the best IDF1 of 74.9\%. BoT-SORT also exhibits improved performance, with IDF1 and IDSw showing enhancements of 1\% and 8.8\%, respectively. These results indicate that our abnormal motion suppression Kalman Filter contributes to better tracking performance across the evaluated trackers.

Next, we apply our pose-guided Re-ID module on all three trackers. The results presented in Table \ref{table: ablation study} demonstrate that the pose-guided Re-ID module primarily impacts the number of IDSw. For JDE, compared to the previous step, the IDSw shows a further improvement of 2.5\%. In FairMOT, the performance on avoiding IDSw improves dramatically by 16.8\%. Similarly, in BoT-SORT, the number of IDSw drops from 134 to 129, indicating a 3.7\% improvement. These findings highlight the efficacy of our pose-guided Re-ID module in reducing IDSw and enhancing tracking performance. 

Finally, we incorporate our occlusion-aware distance measurement into the three trackers. Notably, this component contributes to the IDF1, as observed in all three trackers. The IDF1 shows improvement across the board, with particularly promising results in BoT-SORT, where the IDF1 reached 83.0\%. As for IDSw, improvements are made except for FairMOT. The main reason is that FairMOT adopts a center-based object detector, which is sensitive to IoU distance. Overall, the results demonstrate the effectiveness of our occlusion-aware distance measurement in handling occluded pedestrians and improving tracking performance.

\paragraph{Speed Filter}
We investigate the impact of alternative speed filters, the Gaussian filter, the Laplacian filter, and the Mean Average filter, on our OccluTrack. Each filter is applied to smooth the buffered centering, moving, and scale-changing speeds. The Gaussian filter convolves the buffered speeds with a Gaussian kernel, effectively smoothing the centering, moving speed, and scale-changing speed. The Laplacian filter employs a weighted average of neighboring speeds to smooth the buffered speeds based on their distances. Finally, the Mean Average filter computes the average of all the buffered speeds, as described in Section \ref{subsec:ms}. We summarize the comparisons among these filters based on the MOT17 validation set in Table \ref{table: speed filter}. The results in Table \ref{table: speed filter} indicate that our proposed Mean average filter yields the best performance in terms of IDF1 and IDSw. This filter achieves the highest accuracy in estimating pedestrian moving speeds, leading to improved tracking performance.
\begin{table}[]
    \centering
    \caption{Ablation study on three speed filters over MOT17 validation set.}
    \label{table: speed filter}
    \resizebox{0.35\textwidth}{!}{
    \begin{tabular}{lcc}
    \toprule
        Filter & IDF1(\%)↑ & IDSw↓  \\
        \midrule
        Laplacian & 82.6 & 136 \\ 
        Gaussian  & 82.8 & 126 \\ 
        Mean Average & \textbf{83.0} & \textbf{125} \\ 
        \bottomrule
    \end{tabular}
    }
\end{table}

\paragraph{Feature Embedding Fusion}

We analyzed the impact of different feature embedding fusion methods on the performance of OccluTrack. The evaluated fusion techniques included addition, concatenation, and our proposed adaptive fusion. The results on the MOT17 validation set, presented in Table \ref{table: feature embedding fusion}, indicate that the adaptive fusion method outperforms the addition and concatenation methods, demonstrating its effectiveness in improving tracking performance.

\begin{table}[]
    \centering
    \caption{Ablation study on feature embedding fusion methods over MOT17 validation set.}
    \label{table: feature embedding fusion}
    \resizebox{0.35\textwidth}{!}{
    \begin{tabular}{lcc}
    \toprule
        Fusion & IDF1(\%)↑ & IDSw↓ \\
        \midrule
        Addition & 82.8 & 126 \\ 
        Concatenate  & 82.8 & 126\\ 
        Adaptive Fusion & \textbf{83.0} & \textbf{125} \\ 
        \bottomrule
    \end{tabular}
    }
\end{table}

\paragraph{Abnormal Motion Suppression Parameter}

\begin{figure*}[htbp]
\centering
\begin{minipage}[t]{0.48\linewidth}
\centering
\includegraphics[width=\textwidth]{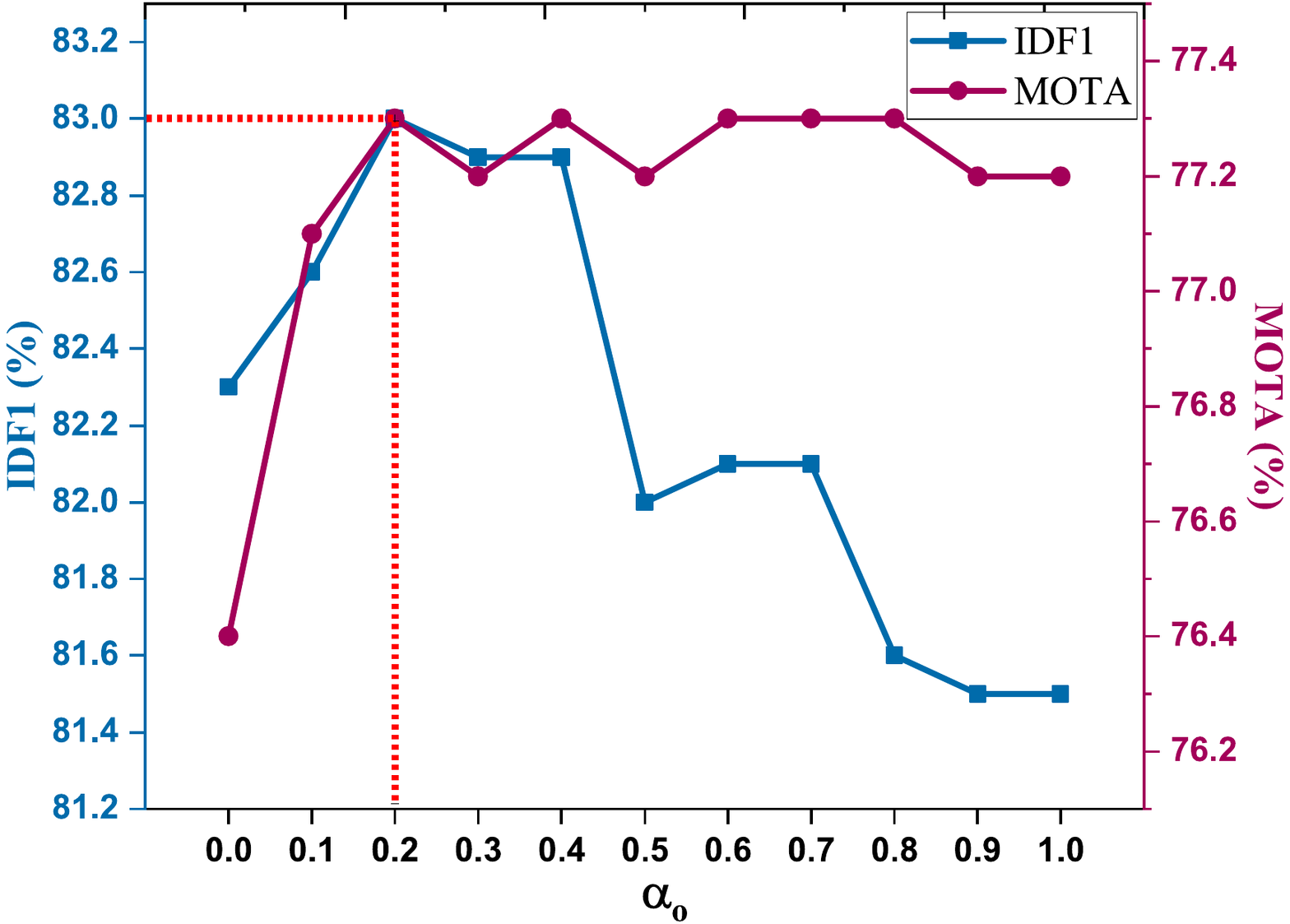}
\caption{Ablation study of abnormal motion suppression parameter $\alpha_0$ on the MOT17 validation set. The $\alpha_0$ is adjusted from 0 to 1. IDF1 and MOTA jointly reach the peak when setting $\alpha_0$=0.2.}
\label{param1}
\end{minipage}%
\hspace{2pt}
\begin{minipage}[t]{0.48\linewidth}
\centering
\includegraphics[width=\textwidth]{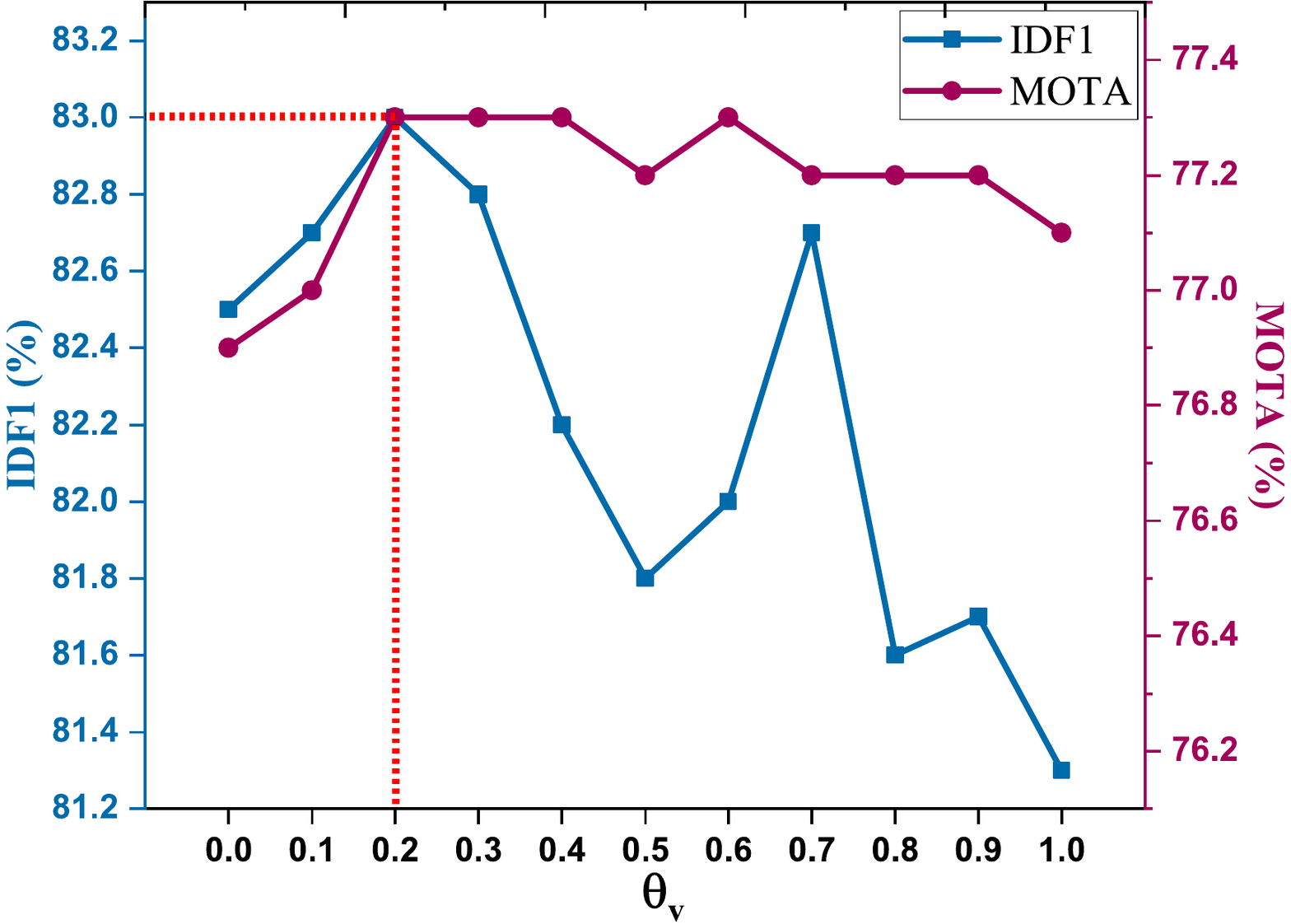}
\caption{Ablation study of abnormal motion detection threshold $\theta_v$ on the MOT17 validation set. The $\theta_v$ is adjusted from 0 to 1. The best IDF1 and MOTA are jointly obtained when setting $\theta_v$=0.2. }
\label{param2}
\end{minipage}
\end{figure*}

In this section, we ablate the abnormal motion suppression parameter $\alpha_0$ on the MOT17 validation set. We vary $\alpha_0$ from 0 to 1 with a step of 0.1 and evaluate its impact on the tracking performance. $\alpha_0=0$ means the Kalman Filter does not update when abnormal motions are detected, while $\alpha_0=1$ means the Kalman Filter does not suppress abnormal motions and operates as the normal Kalman Filter. The results are illustrated in Fig. \ref{param1}. When increasing $\alpha_0$ from 0 to 0.2, we observe that the IDF1 reached the highest at $\alpha_0=0.2$ and MOTA increased from the lowest 76.4\% at $\alpha_0=0$ to the highest 77.3\% at $\alpha_0=0.2$. Further increasing $\alpha_0$ leads to a gradual decline in IDF1 and reaches its lowest 81.5\% at $\alpha_0=1.0$, while MOTA remains relatively stable and ranges between 77.2\% and 77.3\%. Based on these findings, we set $\alpha_0 = 0.2$ for evaluating the MOT17 and MOT20 test sets.

\paragraph{Abnormal Motion Detection Threshold}
Apart from the suppression parameter, we also conduct ablation studies on the effect of adjusting the abnormal motion detection threshold $\theta_v$ from 0 to 1 with a step of 0.1. The results of IDF1 and MOTA are illustrated in Fig. \ref{param2}. When increasing $\theta_v$ from 0 to 0.2, both MOTA and IDF1 achieve the optimal points 77.3 \% and 83.0 \%. With further increasing the threshold, the joint performance of MOTA and IDF1 fluctuates and drops to 77.1 \% and 81.3 \%. In summary, the optimal threshold for abnormal motion detection should be set to 0.2 to obtain the best tracking performance.

\subsection{Benchmark Evaluation}
\label{subsec: benchmark}

We compare our OccluTrack with the state-of-the-art trackers on MOT17 \cite{mot17}, MOT20 \cite{mot20}, and DanceTrack \cite{dancetrack} datasets. We provide two versions of our proposed tracker, OccluTrack and OccluTrack+, building on BoT-SORT \cite{botsort} and BoostTrack++ \cite{boosttrack}, respectively. For MOT17, we benchmark our OccluTrack and OccluTrack+ under private detection protocol and OccluTrack under public protocol in Table \ref{table: MOT17 Benchmark (Private)} and Table \ref{table: MOT17 Benchmark (Public)}. For the private protocol, the detection results are from our object detector, which is trained on multiple datasets. For the public protocol, we filter the YOLOX detection results by calculating the Intersection over Union (IoU) with the detection results provided by MOTChallenge, following ByteTrack \cite{bytetrack}. These results come from detectors such as FRCNN, SDP, and DPM, and we apply an IoU threshold of 0.5 for filtering. For MOT20, we only compare OccluTrack and OccluTrack+ with the state-of-the-art method under the private detection protocols in Table \ref{table: MOT20 Benchmark} as limited published methods provide the results under the public detection protocol. We also provide benchmark comparisons on the DanceTrack dataset under the private detection protocol.

\paragraph{MOT17} Under the private protocol, our OccluTrack excels BoT-SORT on all metrics. Regarding IDF1 and MOTA, our method improves by 1.8\% and 1\% compared with our baseline, which indicates that our method has a more powerful tracking performance while keeping compatible detection performance. Correspondingly, the HOTA improved by 0.7\%, considering both detection and tracking performance. Compared with the most recent work, our OccluTrack outperforms UCMCTrack on IDF1, MOTA, and IDSw by 1.0\%, 0.9\%, and 19.0\%. As for the OccluTrack+, it outperforms existing methods on most metrics like IDF1, HOTA, and IDSw, achieving state-of-the-art performance. Compared with BoostTrack++ (baseline), our OccluTrack+ improves IDF1 and HOTA to 82.8\% (0.6\% ↑) and 66.8\% (0.2\% ↑) while decreasing the ID Switches to 951 (10.5\% ↑). The improvements demonstrating our OccluTrack+ show better tracking performance by mitigating the partial occlusion.

Under public detection, our proposed OccluTrack outperforms existing state-of-the-art methods that provide results under public detection by a large margin. As the results in Table \ref{table: MOT17 Benchmark (Public)} indicate, our methods achieved the best performance among the existing methods on all listed important metrics. Specifically, OccluTrack improves the IDF1, HOTA, MOTA, and IDSw by 7.6\%, 5.2\%, 5.8\%, and 35.3\% compared with ByteTrack \cite{bytetrack}. The improvements show that our OccluTrack has superior tracking performance among existing methods. Noticeably, the proposed tracker can be modified and plugged into any tracking-by-detection method.

\paragraph{MOT20} Even though MOT20 is a more complex dataset with more crowdedness compared with the MOT17 dataset, our OccluTrack and OccluTrack+ still achieve promising performance. Comparing our OccluTrack with BoT-SORT, it achieves better performance in IDF1, HOTA, MOTA, and IDSw. For IDF1 and MOTA, OccluTrack improves them by 1.2\% and 0.1\%, corresponding to tracking performance and detection performance, contributing to 0.8\% improvements on HOTA. This demonstrates that OccluTrack is superior in tracking while keeping good detection performance. As for IDSw, our OccluTrack outperforms our baseline and reduces it by 10.6\% from 1257 to 1124, showing a more stable tracking performance. As for OccluTrack+, it achieves state-of-the-art performance regarding IDF1 and HOTA, improving BoostTrack++ by 0.7\% and 0.3\%, which also decreases IDSw. from 762 to 429. It is evident that our proposed methods are capable of improving tracking performance by considering partial occlusion, especially decreasing the ID Switches in crowded scenes.

\paragraph{DanceTrack}
Apart from MOT17 and MOT20 datasets, we also evaluate our proposed OccluTrack on the DanceTrack dataset to further evaluate the effectiveness of fast-motion scenarios. Different from MOT17 and MOT20, we set the abnormal motion suppression parameters $\alpha_0$ and $\theta_v$ to 0.7, which is less sensitive to fast motion changes. As the results in Table. \ref{table: DanceTrack Benchmark} show, we benchmark our OccluTrack and methods used in MOT17 and MOT20 for fair comparisons. We focus on the three most important and mostly reported metrics: IDF1, HOTA, and MOTA. Our proposed OccluTrack also outperforms existing methods, especially OC-SORT \cite{ocsort}, with noticeable improvements. For IDF1, HOTA, and MOTA, OccluTrack surpasses the OC-SORT \cite{ocsort} by 3.7\%, 1.8\%, and 0.2\% respectively. The improvements of IDF1 and HOTA also demonstrate advanced tracking performance over OC-SORT.

\begin{table}[tp]
    \centering
    
    \caption{Evaluation results and comparisons with the state-of-the-art-trackers on MOT17 test set under the private detection protocol.}
    
    \resizebox{0.5\textwidth}{!}{
        \begin{tabular}{lcccccc}
        \toprule
        Tracker & IDF1↑ & HOTA↑ & MOTA↑ & IDSw↓\\
        \midrule
        
        STTD \cite{sttd} & 51.9 & - & 54.2 & 2735 \\
        QuasiDense \cite{quasi} & 66.3 & 53.9 & 68.7 &   3378  \\ 
        FairMOT \cite{fairmot}  & 72.3 & 59.3 & 73.7 &  3303  \\ 
        TransCenter \cite{transcenter}  & 62.2 & 54.5 & 73.2 &  4614 \\ 
        TransTrack \cite{transtrack}  & 63.5 & 54.1 & 75.2 &  3603   \\ 
        CSTrack \cite{cstrack}   & 72.6 & 59.3 & 74.9  & 3567  \\
        RelationTrack \cite{relationtrack}  & 74.7 & 61.0 & 73.8 &  1374   \\ 
        TransMOT \cite{transmot}  & 75.1 & 61.7 & 76.7 & 2346 \\  
        FDTrack \cite{fdtrack} & 75.6 & 61.3 & 76.8 & 3705  \\
        MOTFR \cite{motfr} & 76.3 & 61.8 & 74.4 & 2625  \\
        OCSORT \cite{ocsort}  & 77.5 & 63.2 & 78.0 &  1950  \\ 
        StrongSORT++ \cite{strongsort}   & 79.5 & 64.4 & 79.6 &  1194 \\ 
        ByteTrack \cite{bytetrack}  & 77.3 & 63.1 & 80.3 & 2196  \\ 
        Quo Vadis \cite{quo-vadis}  & 77.7 & 63.1 & 80.3 &  2103 \\ 
        BoT-SORT \cite{botsort}  & 80.2 & 65.0 & 80.5 &  1212  \\ 
        UCMCTrack \cite{ucmctrack} & 81.0 & 65.7 & 80.6 &1689 \\
        BoostTrack+ \cite{boosttrack} & 81.8 & 66.4 & 80.6 & 1086 \\
        BoostTrack++ \cite{boosttrack} & 82.2 & 66.6 & 80.7 & 1062 \\
        \midrule
        Ours (OccluTrack)  & 82.0 & 65.7 & \textbf{81.5} & 1368\\
        Ours (OccluTrack+) & \textbf{82.8} & \textbf{66.8} & 80.2 & \textbf{951} \\
        \bottomrule
    \end{tabular}
    }
    
\label{table: MOT17 Benchmark (Private)}
\end{table}

\begin{table}[tp]
    \centering
    
    \caption{Evaluation results and comparisons with the state-of-the-art-trackers on MOT17 test set under the public detection protocol.}
    
    \resizebox{0.5\textwidth}{!}{
        \begin{tabular}{lcccccccc}
        \toprule
        Tracker & IDF1↑ & HOTA↑ & MOTA↑ & IDSw↓ \\
        \midrule
        
        Tracktor \cite{tracktor} & 55.1 & 44.8 & 56.3 &  1985 \\
        CenterTrack \cite{centertrack} & 59.6 & 48.2 & 61.5 &  2583 \\
        TMOH \cite{TMOH} & 62.8 & 50.4 & 62.1 & 1897 \\
        QuasiDense \cite{quasi} & 65.1 & - & 64.6 &  2652 \\
        SiamMOT \cite{sotmot} & 63.3 & - & 65.9 & 3040 \\
        Trackformer \cite{trackformer} & 57.6 & - & 62.3 &  4018 \\
        METFormer \cite{metformer} & 59.4 & - & 62.9 & 2827 \\
        ByteTrack \cite{bytetrack} & 70.0 & 56.1 & 67.4 & 1331 \\
        \midrule
        Ours (OccluTrack)  & \textbf{77.6} & \textbf{61.3} & \textbf{73.2} & \textbf{984}\\ 
        \bottomrule
    \end{tabular}
    }
    
\label{table: MOT17 Benchmark (Public)}
\end{table}

\begin{table}[]
    
    \centering
    \caption{Evaluation results and comparisons with the state-of-the-art trackers on MOT20 test set under the private detection protocol.}
    
    \resizebox{0.5\textwidth}{!}{
        \begin{tabular}{lcccc}
        \toprule
        Tracker & IDF1↑ & HOTA↑ & MOTA↑ & IDSw↓  \\ 
        \midrule
        SiamMOT \cite{siammot}  & 69.1 & - & 67.1 & - \\
        FairMOT \cite{fairmot}  & 67.3 & 54.6 & 61.8 &  5243 \\ 
        TransCenter \cite{transcenter}   & 50.4 & - & 61.9 &  5653 \\ 
        TransTrack \cite{transtrack}  & 59.4 & 48.5 & 65.0 &  3608 \\ 
        CSTrack \cite{cstrack}  & 68.6 & 54.0 & 66.6 &  3196 \\ 
        RelationTrack \cite{relationtrack}  & 70.5 & 56.5 & 67.2 &  4243  \\ 
        SOTMOT \cite{sotmot}  & 71.4 & -  & 68.6 &  4209 \\ 
        MOTFR \cite{motfr} & 71.7 & 57.2 & 69.0 & 3648  \\
        FDTrack \cite{fdtrack} & 75.7 & 59.9 & 75.0  & 2226 \\
        OCSORT \cite{ocsort}  & 76.3 & 62.4 & 75.7 &  942 \\ 
        StrongSORT++ \cite{strongsort}   & 77.0 & 62.6 & 73.8 & 770 \\ 
        ByteTrack \cite{bytetrack} & 75.2 & 61.3  & 77.8 & 1223 \\ 
        Quo Vadis \cite{quo-vadis}  & 75.7 & 61.5 & 77.8 & 1185 \\ 
        BoT-SORT \cite{botsort}  & 77.5 & 63.3 & 77.8 & 1257 \\
        UCMCTrack \cite{ucmctrack} &77.4 & 62.8  & 75.6 & 1370 \\
        BoostTrack+ \cite{boosttrack} &81.5 & 66.2 & 77.2 & 827 \\
        BoostTrack++ \cite{boosttrack} &82.0 & 66.4 & 77.7 & 762 \\
        \midrule
         Ours (OccluTrack)  & 78.6 & 64.1 & \textbf{77.9} &  1124 \\ 
         Ours (OccluTrack+) & \textbf{82.7} & \textbf{66.7} & 77.7 & \textbf{429} \\
        \bottomrule
    \end{tabular}
    }
    
\label{table: MOT20 Benchmark}
\end{table}

\begin{table}[tp]
    \centering
    
    \caption{Evaluation results and comparisons with the state-of-the-art trackers on DanceTrack test set.}
    
    \resizebox{1\columnwidth}{!}{
    \renewcommand{\arraystretch}{0.95}
    \small
        \begin{tabular}{lccccccc}
        \toprule
        Tracker & IDF1↑ & HOTA↑ & MOTA↑\\
        \midrule
        CenterTrack \cite{centertrack} & 35.7 & 41.8 & 86.8 \\
        FairMOT \cite{fairmot} & 40.8 & 39.7 & 82.2 \\
        QuasiDense \cite{quasi} & 44.8 & 45.7 & 83.0 \\
        TransTrack \cite{transtrack} & 45.2 & 45.5 & 88.4 \\
        SORT \cite{sort} & 50.8 & 47.9 & 91.8 \\
        DeepSORT \cite{deepsort} & 47.9 & 45.6 & 87.8 \\
        ByteTrack \cite{bytetrack} & 52.5 & 47.3 & 89.5 \\
        OC-SORT \cite{ocsort} & 54.9 & 55.1 & 92.2 \\
        \midrule
        Ours (OccluTrack)  & \textbf{59.6} & \textbf{56.9} & \textbf{92.4} \\ 
        
        \bottomrule
    \end{tabular}
    }
    
\label{table: DanceTrack Benchmark}
\end{table}

\subsection{Visualization}

\begin{figure*}[htp]
  \centering
\includegraphics[width=1\textwidth]{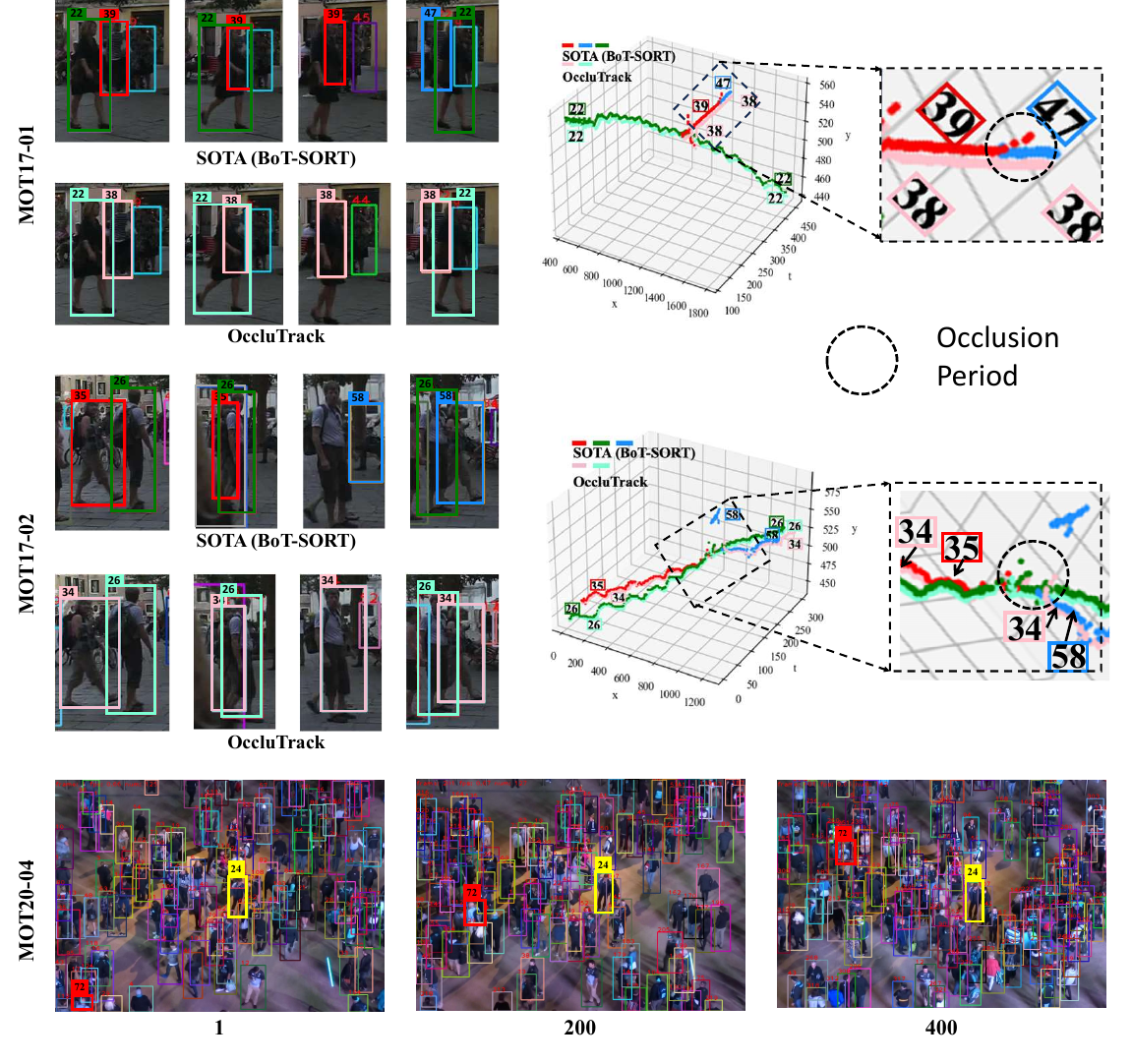}

  \caption{Visualization of our proposed OccluTrack on samples from the MOT17 and MOT20 datasets. On MOT17, we visualize sampled persons and corresponding trajectories with the occlusion period highlighted to show the tracking performance. In MOT-01, our OccluTrack consistently tracks the person with ID 38, while BoT-SORT fails to track the same person with ID 39. A similar situation happens to MOT17-02 for the person with ID 34 in OccluTrack and ID 35 in BoT-SORT.On MOT20, we use the sampled frames in MOT20-04 with two examples (ID 72 and 24) to show that our OccluTrack is also able to handle multiple pedestrian tracking in crowded scenarios.}
  \label{fig:mot_vis}
\end{figure*}

We conducted a comprehensive analysis of tracking results obtained from two videos in the MOT17 and one video from the MOT20, as shown in Fig. \ref{fig:mot_vis}.

On the MOT17 dataset, our purpose is to compare the tracking performance of our OccluTrack with that of the state-of-the-art BoT-SORT under various conditions. We show the results of the two trackers when the selected pedestrians are before partial occlusion, during partial occlusion, during occlusion, and after occlusion. Besides, we demonstrate the trajectories for both trackers on the right of the figure with the occlusion period highlighted. As shown in Fig. \ref{fig:mot_vis} of MOT17-01, BoT-SORT struggled to assign the correct ID to the person with ID 39, whose ID changed to 47 after occlusion. In contrast, our OccluTrack consistently tracked the same person with ID 38 throughout the sequence. This discrepancy in performance can be attributed to the superior accuracy of OccluTrack's predictions during occlusion, as they are closely aligned with the positions of occluded pedestrians. The visualized trajectories further confirm the robust tracking capability of our OccluTrack in maintaining track continuity for the person with ID 38. From the zoomed occlusion period, we can see that BoT-SORT is sensitive to abnormal motions caused by partial occlusion, while our OccluTrack is more robust when occlusion happens. A similar situation happens in MOT17-02. After occlusion, our OccluTrack consistently tracks the person (ID 34) without being affected by partial occlusion, while BoT-SORT fails to track the same person (ID 35). 

On the MOT20 dataset, due to crowded scenes that limit visualization, we selected two examples (ID 24 and 72) in MOT20-04 to showcase the tracking performance of our OccluTrack. We sample three frames from the video at intervals of 200 frames and present the final tracking results in Fig. \ref{fig:mot_vis}. Due to the challenges posed by crowded scenes, it was not feasible to visualize individual trajectories or specific tracks. However, the visualization of example pedestrians (ID 24 and 72), combined with the corresponding numerical results in Section \ref{subsec: benchmark}, shows the impressive tracking performance of our OccluTrack in handling crowded pedestrian tracking within the MOT20 dataset.

\begin{figure}
    \centering
    \includegraphics[width=0.5\textwidth]{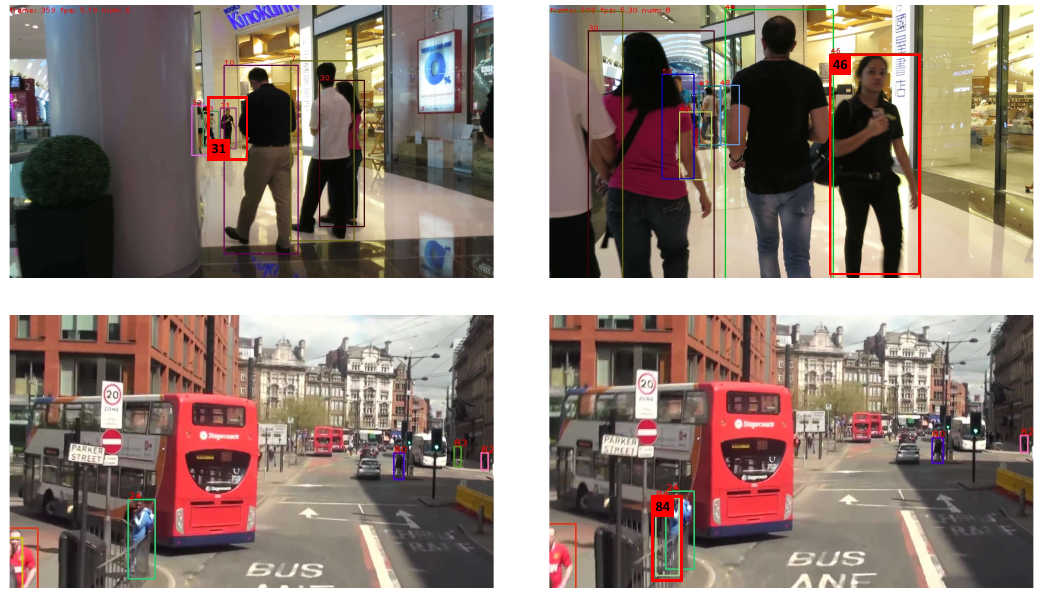}
    \caption{Two examples of typical failure cases. In the first example, when a small-scale pedestrian (ID 31) is occluded in a long period captured by a fast-moving camera, our OccluTrack fails to track the pedestrian and assigns the wrong ID with 46 once it reappears. In the second example, the false negative pedestrian with ID 84 (not exist in the video sequence) is tracked due to the wrong detection results.}
    \label{fig: fail}
\end{figure}

\section{Limitation and Future Work}

\noindent\textbf{Inference Speed with Re-ID. }
The most noticeable limitation of OccluTrack is the slight inference speed drop when compared with our baseline BoT-SORT. Based on the same object detector, the main effects come from two perspectives: pose estimation and the number of pedestrians. For the pose estimation, we compare our OccluTrack with pose-guided Re-ID with the BoT-SORT without pose guidance. With a pose estimator introduced, our current implementation operates at about 7.5 frames per second (FPS) on the MOT17 validation set, while BoT-SORT runs at 8 FPS. However, when the Re-ID module is removed, we can achieve a faster inference speed comparable to methods like BoT-SORT, reaching up to 30 FPS with a 0.4\% drop on IDF1 evaluated on the MOT17 validation set. For the number of pedestrians, we compare the FPS and performance when applying OccluTrack to the MOT17 and MOT20 test sets. MOT17 is constructed with an average pedestrian density of 32, while MOT20 possesses an average pedestrian density of 171. As for the inference speed, the inference speed declined from 7.5 to 1.6 when changing MOT17 to MOT20. We also tested our OccluTrack without pose-guided Re-ID, and the inference speed increased to 7.5 with a tracking performance drop of 0.8 regarding IDF1.

\noindent\textbf{Failure Cases. }
As shown in Fig. \ref{fig: fail}, we highlight two typical failure cases. The first occurs when a small-scale pedestrian is occluded for an extended period while captured by a fast-moving camera. In such cases, the tracking algorithm often struggles to maintain identity due to the challenges posed by the rapid camera movement. For example, the pedestrian with ID 31 in the first example of Fig. \ref{fig: fail}, its ID changes to 46 after such a situation. Specifically, the small scale of the pedestrian, combined with the fast motion of the camera, can cause significant deviations in the pedestrian's trajectory. To mitigate this issue, more advanced techniques, such as optical flow, should be utilized to provide better motion compensation for the fast-moving camera. The second failure case arises from false negatives produced by the object detector. If the detector gives the wrong detections like ID 84 (not exist in the video sequence) in the second example of Fig. \ref{fig: fail}, the tracking algorithm assigns the false negative track. To address this, it is crucial to explore more robust object detectors that can minimize such errors.

\section{Conclusion}
In conclusion, we have shown that abnormal motion caused by partial occlusion is the missing key to enhancing multiple pedestrian tracking performance, and our proposed OccluTrack effectively addresses partial occlusion challenges through three key modules: a plug-and-play abnormal motion suppression mechanism within the Kalman Filter to detect and suppress outlier motions, a pose-guided Re-ID module that captures part-level features of partially visible pedestrians, and an occlusion-aware association strategy for associating occluded pedestrians with a fair distance measurement. Extensive evaluations of MOTChallenge and DanceTrack demonstrate that both OccluTrack and OccluTrack+ outperform the state-of-the-art methods from quantitative and qualitative results. For future work, we aim to optimize the inference speed in crowded scenarios, especially considering the computational overhead introduced by the Re-ID module. In addition, we will explore robust motion modeling and object detection techniques to address failure cases caused by fast-moving cameras and false negative detections.

{
\bibliographystyle{IEEEtran}
\bibliography{References}
}

\vfill

\end{document}